\documentclass[11pt]{article}
\usepackage{colacl}
\usepackage{epsfig} 

\title{
{\Large Learning Correlations between Linguistic Indicators and Semantic
Constraints: }\\  
{\Large Reuse of Context-Dependent Descriptions of Entities}
}

\author{
{\large Dragomir R. Radev} \\
{\large Department of Computer Science} \\
{\large Columbia University} \\
{\large New York, NY 10027} \\
{\large radev@cs.columbia.edu} 
}

\begin{document}
\date{}
\maketitle
\bibliographystyle{acl}

\begin{abstract}
This paper presents the results of a study on the semantic constraints
imposed on lexical choice by certain contextual indicators. We show how
such indicators are computed and how correlations between them and the
choice of a noun phrase description of a named entity can be automatically
established using supervised learning. Based on this correlation, we have
developed a technique for automatic lexical choice of descriptions of
entities in text generation. We discuss the underlying relationship between
the pragmatics of choosing an appropriate description that serves a
specific purpose in the automatically generated text and the semantics of
the description itself. We present our work in the framework of the more
general concept of reuse of linguistic structures that are automatically
extracted from large corpora. We present a formal evaluation of our
approach and we conclude with some thoughts on potential applications of
our method.
\end{abstract}

\thispagestyle{empty}
\pagestyle{empty}

\section{Introduction}

Human writers constantly make deliberate decisions about picking a
particular way of expressing a certain concept. These decisions are made
based on the topic of the text and the effect that the writer wants to
achieve. Such contextual and pragmatic constraints are obvious to
experienced writers who produce context-specific text without much
effort. However, in order for a computer to produce text in a
similar way, either these constraints have to be added manually by an
expert or the system must be able to acquire them in an
automatic way. 


An example related to the lexical choice of an appropriate nominal
description of a person should make the above clear.  Even though it seems
intuitive that Bill Clinton should always be described with the NP {\it
``U. S. president''} or a variation thereof, it turns out that many other
descriptions appear in on-line news stories that characterize him in light
of the topic of the article. For example, an article from 1996 on elections
uses {\it ``Bill Clinton, the democratic presidential candidate''}, while a
1997 article on a false bomb alert in Little Rock, Ark. uses {\it ``Bill
Clinton, an Arkansas native''}.

This paper presents the results of a study of the correlation between named
entities (people, places, or organizations) and noun phrases used to
describe them in a corpus. 

Intuitively, the use of a description is based on a deliberate decision on
the part of the author of a piece of text. A writer is likely to select a
description that puts the entity in the context of the rest of the
article.

It is known that the distribution of words in a document is related
to its topic \cite{Salton&McGill83}. We have developed related techniques
for approximating pragmatic constraints using words that appear in the
immediate context of the entity.

We will show that context influences the choice of a description, as do
several other linguistic indicators.  Each of the indicators by itself
doesn't provide enough empirical data that distinguishes among all
descriptions that are related to a an entity. However, a carefully selected
combination of such indicators provides enough information in order pick
an appropriate description with more than 80\% accuracy.

Section~\ref{section-problem-description} describes how we can
automatically obtain enough constraints on the usage of
descriptions. In Section~\ref{section-language-reuse}, we show how such
constructions are related to language reuse.

In Section~\ref{section-experiment} we describe our experimental setup and
the algorithms that we have designed. Section~\ref{section-results}
includes a description of our results.

In Section~\ref{section-future} we discuss some possible extensions to our
study and we provide some thoughts about possible uses of our framework.

\section{Problem Description}
\label{section-problem-description}

Let's define the relation $DescriptionOf (E)$ to be the one between a named
entity $E$ and a noun phrase, $D$, describing the named entity. In the
example shown in Table~\ref{figure-example1}, there are two
entity-description pairs.

$DescriptionOf$ (``Tareq Aziz'') $=$ ``Iraq's Deputy Prime Minister'' 

$DescriptionOf$ (``Richard Butler'') $=$ ``Chief U.N. arms inspector'' 

\begin{figure}[htbp]
\centering
\small
\begin{tabular}{|l|}
\hline
\\
\underline {{\em Chief U.N. arms inspector} {\bf Richard Butler}}\\
met \underline {{\em Iraq's Deputy Prime Minister} {\bf Tareq Aziz}}\\
Monday after rejecting Iraqi attempts to set \\
deadlines for finishing his work. \\
\\
\hline
\end{tabular}
\caption{Sample sentence containing two entity-description
pairs.} 
\label{figure-example1}
\end{figure}

Each entity appearing in a text can have multiple descriptions (up to
several dozen) associated with it. 

We call the set of all descriptions related to the same entity in a corpus,
a {\it profile} of that entity. Profiles for a large number of entities
were compiled using our earlier system, PROFILE \cite{Radev&McKeown97}. It
turns out that there is a large 
variety in the size of the profile (number of distinct descriptions) for 
different entities. Table~\ref{table-huot-descriptions} shows a subset of
the profile for Ung Huot, the former foreign minister of Cambodia, who
was elected prime minister at some point of time during the run of our
experiment. A few sample semantic features of the descriptions in
Table~\ref{table-huot-descriptions} are shown as separate columns. 

We used information extraction techniques
to collect entities and descriptions from a corpus and analyzed their 
lexical and semantic properties.

We have processed 178 MB\footnote{The corpus contains 19,473 news stories
that cover the period October 1, 1997 - January 9, 1998 that were available
through PROFILE.} of newswire and analyzed the use of descriptions related to
11,504 entities. Even though PROFILE extracts other entities in addition to
people (e.g., places and organizations), we have restricted our analysis to
names of people only. We claim, however, that a large portion of our
findings relate to the other types of entities as well.

We have investigated 35,206 tuples, consisting of an entity, a description,
an article ID, and the position (sentence number) in the article in which
the entity-description pair occurs. Since there are 11,504 distinct
entities, we had on average 3.06 distinct descriptions per entity
($DDPE$). Table~\ref{table-ed} shows the
distribution of $DDPE$ values across the corpus. Notice that a large number
of entities (9,053 out of the 11,504) have a single description. These are
not as interesting for our analysis as the remaining 2,451 entities that
have $DDPE$ values between 2 and 24.

\begin{table*}[htbp]
\centering
\small
\begin{tabular}{|l|c|c|c|c|c|c|}
\hline
& \multicolumn{6}{|c|}{{\bf Semantic categories}} \\ 
\cline{2-7}
{\bf Description} & addressing & country & male & new & political post &
seniority  \\  
\hline
a senior member                      &   &   &   &   &   & X \\  
Cambodia's                           &   & X &   &   &   &   \\  
Cambodian foreign minister           &   & X &   &   & X &   \\  
co-premier                           &   &   &   &   & X &   \\  
first prime minister                 &   &   &   &   & X &   \\  
foreign minister                     &   &   &   &   & X &   \\  
His Excellency                       & X &   &   &   &   &   \\  
Mr.                                  &   &   & X &   &   &   \\  
new co-premier                       &   &   &   & X & X &   \\  
new first prime minister             &   &   &   & X & X &   \\  
newly-appointed first prime minister &   &   &   & X & X &   \\  
premier                              &   &   &   &   & X &   \\  
prime minister                       &   &   &   &   & X &   \\  
\hline
\end{tabular}
\caption{Profile of Ung Huot}
\label{table-huot-descriptions}
\end{table*}

\begin{table*}[htbp]
\centering
\small
\begin{tabular}{||r|r||r|r||r|r||}
\hline
{\bf $DDPE$ } & {\bf count } & {\bf $DDPE$} & {\bf count} & {\bf $DDPE$} &
{\bf count} \\
\hline
1 & 9,053 &  8 & 27 & 15 &  4 \\
2 & 1,481 &  9 & 26 & 16 &  2 \\
3 &  472 & 10 & 12 & 17 &  2 \\
4 &  182 & 11 & 10 & 18 &  1 \\
5 &  112 & 12 &  8 & 19 &  1 \\
6 &   74 & 13 &  2 & 24 &  1 \\
7 &   31 & 14 &  3 &    &    \\
\hline
\end{tabular}
\caption{Number of distinct descriptions per entity ($DDPE$).} 
\label{table-ed}
\end{table*}


\begin{figure}[htbp]
\epsfig{file=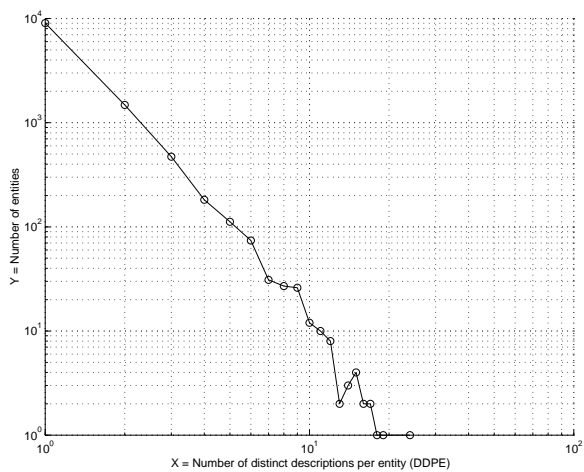,width=3in}
\caption{Number of distinct descriptions per entity (log-log scale)}
\label{figure-ed-log}
\end{figure}

\section{Language Reuse in Text Generation}
\label{section-language-reuse}

Text generation usually involves lexical choice - that is, choosing
one way of referring to an entity over another. Lexical choice refers to a
variety of decisions that have to made in text generation. For example,
picking one among several equivalent (or nearly equivalent) constructions is a form
of lexical choice (e.g., {\it  "The Utah Jazz handed the Boston Celtics a
defeat}" vs. {\it "The Utah Jazz defeated the Boston Celtics"}
\cite{Robin94}). We are interested in a different aspect of the problem:
namely learning the rules that can be used for automatically selecting an
appropriate description of an entity in a specific context.

To be feasible and scaleable, a technique for solving a particular case of
the problem of lexical choice must involve automated learning. It is also
useful if the technique can specify enough constraints on the text to be
generated so that the number of possible surface realizations that match
the semantic constraints is reduced significantly. The easiest case in
which lexical choice can be made is when the full surface structure can be
used, and when it has been automatically extracted from a corpus. Of
course, the constraints on the use of the structure in the generated text
have to be reasonably similar to the ones in the source text.

We have found that a natural application for the analysis of
entity-description pairs is {\it language reuse}, which includes techniques
of extracting shallow structure from a corpus and applying that structure
to computer-generated texts.

Language reuse involves two components: a {\it source} text written by a
human and a {\it target} text, that is to be automatically generated by a
computer, partially making use of structures reused from the {\it source}
text. The {\it source} text is the one from which particular surface 
structures are extracted automatically, along with the appropriate 
syntactic, semantic, and pragmatic constraints under which they are used. 

Some examples of language reuse include collocation analysis 
\cite{Smadja93}, the use of entire factual
sentences extracted from corpora (e.g., {\it ``'Toy Story' is the Academy
Award winning animated film developed by Pixar}''), and summarization using
sentence extraction \cite{Paice90,Kupiec&al95}. In the case
of summarization through sentence extraction, the target text has the
additional property of being a subtext of the source text. Other techniques
that can be broadly categorized as language reuse are learning
relations from on-line texts \cite{Mitchell97} and answering natural
language questions using an on-line encyclopedia \cite{Kupiec93}. 

Stydying the concept of language reuse is rewarding because it allows
generation systems to leverage on texts written by humans and their
deliberate choice of words, facts, structure.

We mentioned that for language reuse to take place, the generation system
has to use the same surface structure in the same syntactic, semantic, and
pragmatic context as the source text from which it was
extracted. Obviously, all of this information is typically not available to
a generation system. There are some special cases in which most of it can
be automatically computed.

Descriptions of entities are a particular instance of a surface structure
that can be reused relatively easily. {\it Syntactic} constraints related
to the use of descriptions are modest - since descriptions are always noun
phrases that appear as either pre-modifiers or appositions\footnote{We
haven't included relative clauses in our study.}, they are quite flexibly
usable in any generated text in which an entity can be modified with an
appropriate description. We will show in the rest of the paper how the
requisite {\it semantic} (i.e., {\it "what is the meaning of the
description to pick"}) and {\it pragmatic} constraints (i.e., {\it "what
purpose does using the description achieve?"}) can be extracted
automatically. 

Given a profile like the one shown in Table~\ref{table-huot-descriptions},
and an appropriate set of semantic constraints (columns 2-7 of the table),
the generation component needs to perform a profile lookup and select a
row (description) that satisfies most or all semantic constraints. For
example, if the semantic constraints specify that the description has to
include the country and the political position of Ung Huot, the most
appropriate description is {\it "Cambodian foreign minister"}. 



\section{Experimental Setup}
\label{section-experiment}

In our experiments, we have used two widely available tools - 
WordNet and Ripper.

WordNet \cite{Miller&al90} is an on-line hierarchical lexical database which
contains semantic information about English words (including hypernymy
relations which we use in our system). We use chains of hypernyms when we
need to approximate the usage of a particular word in a description using
its ancestor and sibling nodes in WordNet. Particularly useful for our
application are the {\it synset offsets} of the words in a description. The
{\it synset offset} is a number that uniquely identifies a concept node ({\it
synset}) in the WordNet hierarchy. Figure~\ref{figure-wordnet} shows that
the synset offset for the concept {\it ``administrator, decision maker''}
is {\it ``\{07063507\}''}, while its hypernym, {\it ``head, chief, top
dog''} has a synset offset of {\it ``\{07311393\}''}.

\begin{figure*}[htbp]
\centering
\small
\begin{tabular}{|llllllll|}
\hline
 & & & & & & & \\
 {\bf DIRECTOR:} & \multicolumn{6}{l}{\{07063762\} director, manager, managing
  director} & \\ 
 & \multicolumn{6}{l}{$\Rightarrow$ \{07063507\} administrator, decision
    maker} & \\ 
 & & \multicolumn{5}{l}{$\Rightarrow$ \{07311393\} head, chief, top dog} & \\
 & & & \multicolumn{4}{l}{$\Rightarrow$ \{06950891\} leader} & \\
 & & & & \multicolumn{3}{l}{$\Rightarrow$ \{00004123\} person, individual,
          someone, somebody, mortal, human, soul} & \\ 
 & & & & & \multicolumn{2}{l}{$\Rightarrow$ \{00002086\} life form, organism,
            being, living thing} & \\ 
 & & & & & & \multicolumn{1}{l}{$\Rightarrow$ \{00001740\} entity,
              something} & \\   
 & & & & & & & \\
\hline
\end{tabular}
\caption{Hypernym chain of ``director'' in WordNet, showing synset offsets.} 
\label{figure-wordnet}
\end{figure*}

Ripper \cite{Cohen95} is an algorithm that learns rules from
example tuples in a relation. Attributes in the tuples can be integers
(e.g., length of an article, in words), sets (e.g., semantic features), or
bags (e.g., words that appear in a sentence or document).  We use Ripper to
learn rules that correlate context and other linguistic indicators with the
semantics of the description being extracted and subsequently reused. It is
important to notice that Ripper is designed to learn rules that classify
data into atomic classes (e.g., {\it ``good''}, {\it ``average''}, and {\it
``bad''}). We had to modify its algorithm in order to classify data into
{\it sets of atoms}. For example, a rule can have the form {\it ``if
CONDITION then [\{07063762\} \{02864326\} \{00017954\}]''}\footnote{These
offsets correspond to the WordNet nodes {\it ``manager''}, {\it
``internet''}, and {\it ``group''}}. This rule states that if a certain
``CONDITION'' (which is a function of the indicators related to the
description) is met, then the description is likely to contain words that
are semantically related to the three WordNet nodes [\{07063762\}
\{02864326\} \{00017954\}].

The stages of our experiments are described in detail in the remainder of
this section. 

\subsection{Semantic tagging of descriptions}

Our system, PROFILE, processes WWW-accessible newswire on a round-the-clock basis
and extracts entities (people, places, and organizations) along with
related descriptions.  The extraction grammar, developed in CREP  
\cite{Duford93}, covers a variety of pre-modifier and appositional noun
phrases. 

For each word $w_i$ in a description, we use a version of WordNet to
extract the {\it synset offset} of the immediate parent of $w_i$. 

\subsection{Finding linguistic cues}

Initially, we were interested in discovering rules manually and then
validating them using the learning algorithm. However, the task proved
(nearly) impossible considering the sheer size of the corpus. One possible
rule that we hypothesized and wanted to verify empirically at this stage
was {\it parallelism}. This linguistically-motivated rule states that in a
sentence with a parallel structure (consider, for instance, the sentence
fragment {\it "... Alija Izetbegovic, a Muslim, Kresimir Zubak, a Croat,
and Momcilo Krajisnik, a Serb..."}) all entities involved have similar
descriptions. However, rules at such a detailed syntactic level take too
long to process on a 180 MB corpus and, further, no more than a handful
of such rules can be discovered manually.  As a result, we made a decision to
extract all indicators automatically. We would also like to note that using
syntactic information on such a large corpus doesn't appear particularly
feasible. We limited therefore our investigation to lexical, semantic, and
contextual indicators only. The following subsection describes the
attributes used.

\subsection{Extracting linguistic cues automatically}

The list of indicators that we use in our system are the following:

\begin{itemize}
\item {\bf Context:} (using a window of size 4, excluding the actual
description used, but not the entity itself) - e.g., {\it ``['clinton'
'clinton' 'counsel' 'counsel' 'decision' 'decision' 'gore' 'gore' 'ind'
'ind' 'index' 'news' 'november' 'wednesday' ]''} is a bag of words found
near the description of Bill Clinton in the training corpus.
\item {\bf Length of the article:} - an integer.
\item {\bf Name of the entity:} - e.g., ``Bill Clinton''.
\item {\bf Profile:} The entire profile related to a person (all
descriptions of that person that are found in the training corpus).
\item {\bf Synset Offsets:} - the WordNet node numbers of all words (and
their parents)) that appear in the profile associated with the entity that
we want to describe. 
\end{itemize}

\subsection{Applying machine learning method}

To learn rules, we ran Ripper on 90\% (10,353) of the entities in the entire
corpus. We kept the remaining 10\% (or 1,151 entities) for evaluation.

Sample rules discovered by the system are shown in Table~\ref{table-ripper}.

\begin{table*}[htbp]
\centering
\small
\begin{tabular}{|l|l|}
\hline
{\bf Rule} & {\bf Decision} \\
\hline
IF CONTEXT \~{} inflation & \{09613349\} (politician) \\
IF PROFILES \~{} detective AND CONTEXT \~{ } agency & \{07485319\} (policeman) \\
IF CONTEXT \~{} celine & \{07032298\} (north\_american) \\
\hline
\end{tabular}
\caption{Sample rules discovered by the system.}
\label{table-ripper}
\end{table*}

\section{Results and Evaluation}
\label{section-results}

We have performed a standard evaluation of the precision and recall that
our system achieves in selecting a description. Table~\ref{table-evaluation} 
shows our results under two sets of parameters.

Precision and recall are based on how well the system predicts a set of
semantic constraints. Precision (or $P$) is defined to be the number of
matches divided by the number of elements in the predicted set. Recall (or
$R$) is the number of matches divided by the number of elements in the
correct set. If, for example, the system predicts {\it [A] [B] [C]}, but
the set of constraints on the actual description is {\it [B] [D]}, we would
compute that $P = 33.3\%$ and $R = 50.0\%$. Table~\ref{table-evaluation}
reports the average values of $P$ and $R$ for all training
examples\footnote{We run Ripper in a so-called ``noise-free mode'', which
causes the condition parts of the rules it discovers to be mutually
exclusive and therefore, the values of $P$ and $R$ on the training data are
both 100\%.}.

\begin{table*}[htbp]
\centering
\small
\begin{tabular}{||r||r|r||r|r||}
\hline
 & \multicolumn{2}{r||}{{\bf word nodes only}} 
   & \multicolumn{2}{r||}{{\bf word and parent nodes}} \\
\hline
{\bf Training set size} & {\bf Precision} & {\bf Recall} & {\bf Precision} &
 {\bf Recall} \\  
\hline
500    & 64.29\% & 2.86\%  & 78.57\% &  2.86\% \\
1,000  & 71.43\% & 2.86\%  & 85.71\% &  2.86\% \\
2,000  & 42.86\% & 40.71\% & 67.86\% & 62.14\% \\
5,000  & 59.33\% & 48.40\% & 64.67\% & 53.73\% \\
10,000 & 69.72\% & 45.04\% & 74.44\% & 59.32\% \\
15,000 & 76.24\% & 44.02\% & 73.39\% & 53.17\% \\
20,000 & 76.25\% & 49.91\% & 79.08\% & 58.70\% \\
25,000 & 83.37\% & 52.26\% & 82.39\% & 57.49\% \\
30,000 & 80.14\% & 50.55\% & 82.77\% & 57.66\% \\
50,000 & 83.13\% & 58.54\% & 88.87\% & 63.39\% \\
\hline
\end{tabular}
\caption{Values for precision and recall using word nodes only (left) and
both word and parent nodes (right).}
\label{table-evaluation}
\end{table*}

Selecting appropriate descriptions based on our algorithm is feasible
even though the values of precision and recall obtained may seem only
moderately high. The reason for this is that the problem that we are trying
to solve is underspecified. That is, in the same context, more than one
description can be potentially used. Mutually interchangeable descriptions
include synonyms and near synonyms ({\it ``leader''} vs. {\it ``chief}) or
pairs of descriptions of different generality ({\it U.S. president}
vs. {\it president}). This type of evaluation requires the availability of
human judges.

There are two parts to the evaluation: how well does the system performs in
selecting semantic features (WordNet nodes) and how well it works in
constraining the choice of a description. To select a description, our
system does a lookup in the profile for a possible description that
satisfies most semantic constraints (e.g., we select a row in
Table~\ref{table-huot-descriptions} based on constraints on the columns). 

Our system depends crucially on the multiple components that we use. For
example, the shallow CREP grammar that is used in extracting entities and 
descriptions often fails to extract good descriptions, mostly due to
incorrect PP attachment. We have also had problems from the part-of-speech
tagger and, as a result, we occasionally incorrectly extract word
sequences that do not represent descriptions.

\section{Applications and Future Work}
\label{section-future}

We should note that PROFILE is part of a large system for
information retrieval and summarization of news through information
extraction and symbolic text generation \cite{McKeown&Radev95}. We intend
to use PROFILE to improve lexical choice in the summary generation component,
especially when producing user-centered summaries or summary updates
\cite{Radev&McKeown98}. There are two particularly appealing cases - (1)
when the extraction component has failed to extract a description and (2)
when the user model (user's interests, knowledge of the entity and personal
preferences for sources of information and for either conciseness or
verbosity) dictates that a description should be used even when one doesn't
appear in the texts being summarized.

A second potentially interesting application involves using the data and
rules extracted by PROFILE for {\it language regeneration}. In
\cite{Radev&McKeown98} we show how the conversion of extracted
descriptions into components of a generation grammar allows for flexible
(re)generation of new descriptions that don't appear in the source
text. For example, a description can be replaced by a more general one, two
descriptions can be combined to form a single one, or one long description
can be deconstructed into its components, some of which can be reused as
new descriptions.

We are also interested in investigating another idea - that of predicting
the use of a description of an entity even when the corresponding profile
doesn't contain any description at all, or when it contains only
descriptions that contain words that are not directly related to the words
predicted by the rules of PROFILE. In this case, if the system predicts a  
semantic category that doesn't match any of the descriptions in a specific
profile, two things can be done: (1) if there is a single description in
the profile, to pick that one, and (2) if there is more than one
description, pick the one whose semantic vector is closest to the predicted
semantic vector. 

Finally, the profile extractor will be used as part of a large-scale,
automatically generated {\it Who's who} site which will be
accessible both by users through a Web interface and by NLP systems through
a client-server API. 

\section{Conclusion}
\label{section-conclusion}

In this paper, we showed that context and other linguistic indicators
correlate with the choice of a particular noun phrase to describe an
entity. Using machine learning techniques from a very large corpus, we 
automatically extracted a large set of rules that predict the choice of a
description out of an entity profile. We showed that high-precision
automatic prediction of an appropriate description in a specific context is
possible. 

\section{Acknowledgments}
\label{section-acknowledgments}

This material is based upon work supported by the National Science
Foundation under Grants No. IRI-96-19124, IRI-96-18797, and CDA-96-25374,
as well as a grant from Columbia University's Strategic Initiative Fund
sponsored by the Provost's Office. Any opinions, findings, and conclusions
or recommendations expressed in this material are those of the author(s) and
do not necessarily reflect the views of the National Science Foundation.

The author is grateful to the following people for their comments and
suggestions: Kathy McKeown, Vasileios Hatzivassiloglou, and Hongyan Jing. 

\bibliography{reuse}

\end{document}